\documentclass[sigconf]{acmart}

\usepackage{stfloats} %adjust the layout of figures

\usepackage{booktabs} % For formal tables
\copyrightyear{2017} 
\acmYear{2017} 
\setcopyright{acmlicensed}
\acmConference{CIKM'17}{}{November 6--10, 2017, Singapore.}
\acmPrice{15.00}
\acmDOI{http://dx.doi.org/10.1145/3132847.3133169}
\acmISBN{ISBN 978-1-4503-4918-5/17/11} 
 
\begin{document}
\title{\emph{AliMe Assist}: An Intelligent Assistant for Creating an Innovative E-commerce Experience}
%\titlenote{Produces the permission block, and copyright information}
%\subtitle{Extended Abstract}
%\subtitlenote{The full version of the author's guide is available as \texttt{acmart.pdf} document}

\author{Feng-Lin Li, Minghui Qiu, Haiqing Chen, Xiongwei Wang, Xing Gao, Jun Huang, Juwei Ren, Zhongzhou Zhao, Weipeng Zhao, Lei Wang, Guwei Jin, Wei Chu}
\affiliation{\institution{Alibaba Group}}
\email{fenglin.lfl@alibaba-inc.com}

%\author{Anonymous Authors}
%\affiliation{\institution{Alibaba Group}}
%\email{xxx@alibaba-inc.com}

\renewcommand{\shortauthors}{F-L. Li et al.}
%\renewcommand{\shortauthors}{Anonymous Authors}

%Through

\begin{abstract}
We present \emph{AliMe Assist}, an intelligent assistant designed for creating an innovative online shopping experience in E-commerce. Based on \emph{question answering} (QA), \emph{AliMe Assist} offers assistance service, customer service, and chatting service. It is able to take voice and text input, incorporate context to QA, and support multi-round interaction. Currently, it serves millions of customer questions per day and is able to address 85\% of them. In this paper, we demonstrate the system, present the underlying techniques, and share our experience in dealing with real-world QA in the E-commerce field.
\end{abstract}
%(include shopping guidance and after-sales service)
%We present \emph{AliMe Assist}, an intelligent assistant designed for creating an innovative online shopping experience in E-commerce

%\begin{CCSXML}
%<ccs2012>
%<concept>
%<concept_id>10010147.10010178.10010179.10010181</concept_id>
%<concept_desc>Computing methodologies~Discourse, dialogue and pragmatics</concept_desc>
%<concept_significance>500</concept_significance>
%</concept>
%<concept>
%<concept_id>10010147.10010178.10010179.10010182</concept_id>
%<concept_desc>Computing methodologies~Natural language generation</concept_desc>
%<concept_significance>300</concept_significance>
%</concept>
%</ccs2012>
%\end{CCSXML}

%\ccsdesc[500]{Computing methodologies~Discourse, dialogue and pragmatics}
%\ccsdesc[300]{Computing methodologies~Natural language generation}


\maketitle

\section{Introduction}
%a headline technology 
%substantial
%achieved
During the last few years, \emph{question answering} (QA) based intelligent assistants have been very popular -- partly due to the progresses achieved in deep learning and big data techniques, and partly due to the growing requirements in the real-world.

%Customer service is one of the promising fields that intelligent assistants can play a key role in. On one hand, with the fast-growing E-commerce (EC) markets~\footnote{According to the 2016 (first half) Annual Data Monitoring Report, the GMV of China EC market was 10.5 trillion with an increase of 37.6\% year-on-year.}, there is a sharp increase in demand for customer service staff. Obviously, keep hiring is impractical since the cost will constantly grow, and, the continuous expansion may turn many EC companies into customer service centers. On the other hand, with consumption upgrade, nowadays more and more customers attach importance to shopping experience and service quality. However, traditional customer service suffers from serious issues, e.g., limited availability (only office hours), inefficiency (customers often have to wait minutes), and long turn-on time, especially during peak period (e.g., the ``Double 11'' day). 

%Obviously, keep hiring is impractical since the cost will constantly grow, and, the continuous expansion may turn many EC companies into customer service centers.
%a sharp increase in demand for

Customer service is one of the promising fields that intelligent assistants can play a key role in. On one hand, there is a strong demand for customer service staff in the E-commerce (EC) filed along with the fast-growing EC market. For example, according to the 2016 (first half) Annual Data Monitoring Report, the GMV of China EC market was 10.5 trillion with an increase of 37.6\% year-on-year. On the other hand, with more and more people paying attention to shopping experience and service quality nowadays, the issues of traditional customer service are becoming obvious, e.g., limited availability (only office hours), inefficiency (customers often have to wait minutes), and long turn-on time, especially during peak period (e.g., the ``Double 11'' day).

%However, hiring more and more staff is impractical since the cost will keep growing, and, the continuous expansion may turn many EC companies into customer service centers.

% the ``Double 11'' day 
% promotion days
%This raises a question: is it possible to design an intelligent customer service assistant that could relieve customer service staff from addressing simple, common and repetitive questions, and let them focus on the cases that really need human participation? If so, we can largely facilitate productivity, improve customer experience and substantially reduce labor cost.
%addressing
%customer service

This raises an important question: is it possible to design an intelligent assistant that can relieve customer service staff from answering simple, common and repetitive questions, and let them focus on the cases that really need human participation? If so, we can largely facilitate productivity, improve user experience and reduce cost. %labor

%Along this direction, we have been working on intelligent customer service assistant in the EC field for years. Our assistant, \emph{AliMe Assist}, is able to offer three kinds of service: operation assistance, after-sales service and chatting service. Moreover, it is able to take voice, text and image input, incorporate context to QA, and support multi-round session interaction. Currently, it serves millions of customer questions per day (mainly Chinese, also English) and is able to address 85\% of them. % session

We have been working on this topic for years. Our intelligent assistant, \emph{AliMe Assist}, offers three kinds of services: assistance service, customer service and chatting service. Moreover, it is able to take voice and text input, incorporate context to QA, and support multi-round interaction. Currently, it serves millions of customer questions per day (mainly Chinese, also some English) and is able to address 85\% of them. In this paper, we present the system and the underlying techniques, and share our experience in dealing with real-world QA scenarios in the E-commerce industry. 

%present the approaches proposed in support of the features

This work makes the following contributions:
%This work makes the following contributions:
%which determines the dispatching of questions to customer service staff group (if needed)
%multi-classification

\begin{itemize}
\item Designs and develops a real-world industrial intelligent assistant that offers customers with assistance service, customer service and chatting service in E-commerce.
\item Presents a \emph{Convolutional Neural Network} (CNN) model that incorporates the context of customer questions for intention identification.
\item Proposes a \emph{semantic normalization} and \emph{knowledge graph} based approach for knowledge-oriented customer service question answering. 
\item Proposes a hybrid approach that uses an attentive \emph{Sequence-to-Sequence} model to optimize the joint results of information retrieval and generation model for chatting.
\end{itemize}

%\emph{Sequence-to-Sequence} (Seq2Seq)
%Proposes a \emph{trie} based pattern match and \emph{knowledge graph} based query approach for knowledge-oriented customer service question answering.
%\item Proposes a multi-classification DNN model that incorporates context into customer questions for intention identification, which determines the routing of questions among the services (and subordinate scenarios).
%\item Proposes a hybrid approach that combine knowledge graph and information retrieval for after-sales service (knowledge-oriented QA).
%\item Proposes a hybrid approach that makes use of information retrieval and sequence-to-sequence generation for chatting.
%\item We have implemented these three approaches in an intelligent customer service assistant and released an official product.
%\item We propose different approaches for the three kinds of services: {slot filling} for operation assistance, {knowledge graph} for after-sales service, and an attentive sequence-to-sequence (Seq2Seq) based rerank model for chatting. 
%\item We propose a slot filling approach for task-oriented QA.
%which coordinates the services and associated scenarios.

%The rest of the paper is structured as follows: Section~\ref{sec:arch_flow} presents an overview of the system; Section~\ref{sec:approach} discusses the offered services and proposed approaches; Section~\ref{sec:relwork} reviews related work, and Section~\ref{sec:conclusion} concludes the paper and sketches directions for future work.

The rest of the paper is structured as follows: Section~\ref{sec:arch_flow} presents an overview of the system; Section~\ref{sec:approach} discusses the system features and the underlying techniques; Section~\ref{sec:tool} gives a demonstration of the system; Section~\ref{sec:relwork} reviews related work, and Section~\ref{sec:conclusion} concludes the paper and sketches directions for future work.

\section{System Overview}
\label{sec:arch_flow}
%\emph{AliMe Assist} is designed for creating an innovative retail experience for customers. It is built on mass consumer data, and offers after-sales service, operation assistance and chitchat in the form of ``AI (Machine Intelligence) + Human''. 
%online shopping guidance, 

%\emph{AliMe Assist} is designed for creating an innovative online shopping experience for customers. Specifically, it offers three kinds of services, namely operation assistance, after-sales service and chatting service,  in the form of ``AI + Human''. 

%\subsection{Overall Architecture}
%\label{sec:arch}
%~\footnote{The architecture and process flow of \emph{AliMe Assist} are simplified in this paper due to space limitation. For example, the \emph{recommendation engine}  is omitted.}

\begin{figure}[ht]
    \centering
    %\vspace{-0.2cm}
    \includegraphics[width=0.9\columnwidth]{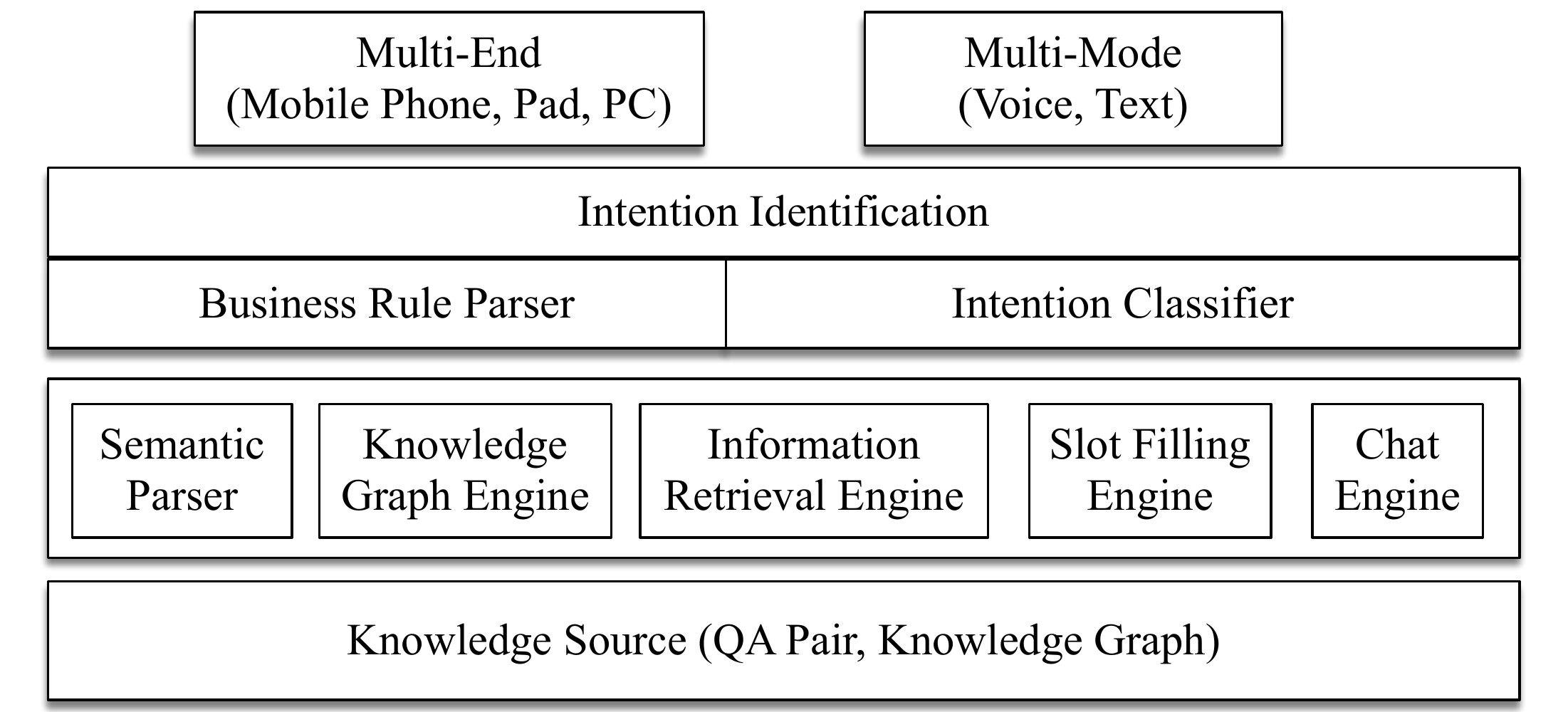}
    \vspace{-0.3cm}
    \caption{The overall architecture of \emph{AliMe Assist}.}
    \label{fig:architecture}
    \vspace{-0.35cm}
 \end{figure}

% which specific scenario?
%  in a layered manner 
We show the architecture of \emph{AliMe Assist} in Fig.~\ref{fig:architecture}. The first layer is the input layer, which supports voice and text input from multi-end (e.g., mobile phone, pad, PC); the second layer sketches the intention layer, which determines the routing of each question  (e.g., assistance service or customer service?); the third layer illustrates the components used for processing questions; and, finally, the forth layer stands for the knowledge source (QA pairs and knowledge graph), from which answers are retrieved. 

%\subsection{Processing Flow}
%\label{sec:flow}
%which consists of a set of patterns and a \emph{trie} for matching the input to potential patterns

%We show the processing flow of customer questions in \emph{AliMe Assist} in Fig.~\ref{fig:processing_flow}. Given an input question $q$, it first passes through the \emph{business rule parser}, a \emph{trie}-based pattern matcher. If $q$ matches certain pattern(s), it will be further judged: if it asks for online customer service, e.g., ``呼唤真人(real person, please)'', it will be directed to staff service; if it is about promotional activities, e.g., ``红包入口在哪里? (what is the entry of grabbing red envelopes?)'', a pre-configured answer will be returned; if it is about assistance service, e.g., ``我想订机票(I want to book a flight ticket)'', it will be taken as task-oriented and handled by a \emph{slot filling engine} (Section~\ref{sec:task_qa}). 

\begin{figure}[h]
    \centering
    \vspace{-0.2cm}
    \includegraphics[width=0.95\columnwidth]{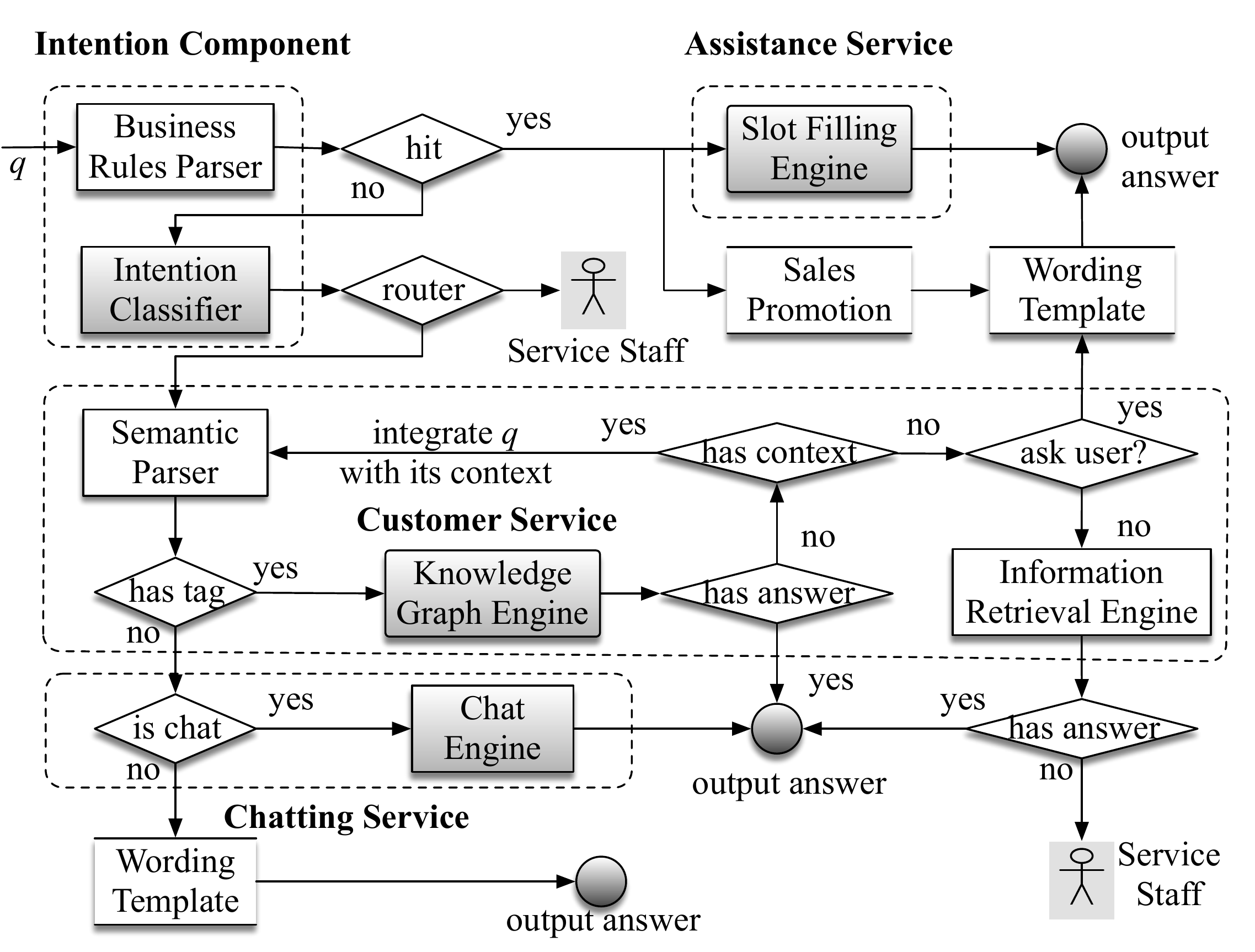} %processing_flow_s
    \vspace{-0.3cm}
    \caption{The overall processing flow of \emph{AliMe Assist}}
    \label{fig:processing_flow}
    \vspace{-0.25cm}
%\end{figure*}
\end{figure}

%(Section~\ref{sec:task_qa})
%(Section~\ref{sec:intention})
We show the processing flow of customer questions in \emph{AliMe Assist} in Fig.~\ref{fig:processing_flow}. Given an input question $q$, it first passes through the \emph{business rule parser}, a \emph{trie}-based pattern matcher. If $q$ matches certain pattern(s), it will be judged as follows: if it requests for task-oriented assistance service, e.g., ``I want to book a flight ticket'', it will be handled by a \emph{slot filling engine}; if it asks about promotional activities, e.g., ``what is the entry of grabbing red envelopes?'', a pre-configured answer will be returned; if it is for online service staff, e.g., ``real person, please'', \emph{AliMe Assist} will ask the user to provide a description of her/his problem. If $q$ does not match any pattern, it will be sent to an \emph{intention classifier} for classification. That is, to be labelled with intention scenarios such as ``return of sales'' and ``refund'', depend on which $q$ will be directed to different service staff group if human participation is needed.

%\begin{figure*}[ht]

%We show the overall processing flow of customer questions in \emph{AliMe Assist} in Fig.~\ref{fig:processing_flow}. Given an input question $q$, it first passes through the \emph{business rule parser}, a \emph{trie}-based pattern matcher. If $q$ matches certain pattern(s), it will be further judged: if it is a request for online customer service (e.g., ``real person, please!''), it will be directed to staff service; if it is a question about promotional activities (e.g., ``what is the entry of grabbing red packets?''), a pre-set answer will be returned; it it requires operation assistance (e.g., ``I want to book a flight ticket''), it will be taken as a task-oriented QA and will be processed as in Section~\ref{sec:task_qa}. 

%If $q$ does not match any pattern, it will be sent to an \emph{intention classifier} for classification (i.e., to be tagged with scenarios such as ``return of sales'' and ``refund'', Section~\ref{sec:intention}), and fed into a \emph{trie}-based \emph{semantic parser}. If any semantic tag (entities in knowledge graph, e.g., ``customer account'') is identified, $q$ will be treated as business related (knowledge-oriented), and the identified tags will be used to retrieve answer (knowledge) through a \emph{knowledge graph engine}. If no answer is retrieved, \emph{AliMe Assist} will enrich $q$ with its context (the previous question) and sent the concatenation back to the \emph{semantic parser} again. 

Next, $q$ will be fed into a \emph{trie}-based \emph{semantic parser}. If any semantic tag (entity in knowledge graph, e.g., ``customer account'') is identified, $q$ will be treated as business related (knowledge-oriented), and the identified tags will be used to retrieve answer through a \emph{knowledge graph engine}. If no answer is retrieved, \emph{AliMe Assist} will enrich $q$ with its context (the previous question) and sent the concatenation back to the \emph{semantic parser} again. 

%(knowledge)
% for semantic tag parsing
%(i.e., to be tagged with scenarios such as ``return of sales'' and ``refund'', Section~\ref{sec:intention})
%If $q$ does not match any pattern, it will be sent to an \emph{intention classifier} for classification (i.e., be tagged with scenarios such as ``customer account'', ``return of sales'' and ``refund''), and fed into a \emph{semantic parser} for semantic tag parsing. If any semantic tag is identified, $q$ will be treated as business related (knowledge-oriented), and the identified tags will be used to retrieve answer (knowledge) through a \emph{knowledge graph engine}. If no answer is retrieved, \emph{AliMe Assist} will enrich $q$ with its context (the previous question) and sent the concatenation back to the \emph{semantic parser} again. 

%$q$ does not have a previous question when it starts a session
% and decide whether or not to ask back to customers
%There are two points to be noted. First, the question $q$ may do not have a context (i.e., $q$ is the first question of a session). Second, it could be the case that there is still no answer even with the concatenation. In both cases, \emph{AliMe Assist} will move to the next step (to avoid endless loop, it uses the context of $q$ only once, if exists): asking customers to provide more information if the identified tags include only one entity/action, otherwise passing on $q$ to the \emph{retrieval engine}. 

There are two points to be noted. First, the question $q$ may not have a context (i.e., $q$ is the first question of a session). Second, it could be the case that there is still no answer even with the concatenation. In both cases, \emph{AliMe Assist} will move to the next step (i.e., it uses the context of $q$ only once, if exists): asking customers for more information if the identified tags include only one entity or action, otherwise passing on $q$ to the \emph{retrieval engine}. In case the retrieval model still fails to give an answer, $q$ will be transferred to service staff according to the labelled intention scenario (our \emph{recommendation engine} is omitted here due to space limitation).

%In the case that
%~\footnote{We omit the \emph{recommendation engine} that follows the retrieval model here.}.

%(customer service staffs are grouped according to scenarios)
%~\footnote{In fact, we have a \emph{recommendation engine} after retrieval model, we omit it in this paper because of space limitation.}.
%On the other hand, if the question $q$ is irrelevant to business and is taken as a chat, it will be fed into a \emph{chat engine}. If $q$ is not a chat, it will be further judged against a rubbish corpus: a pre-configured answer will be outputted (if it is a rubbish), or the \emph{retrieval engine} will be employed (if not).

On the other hand, if $q$ is irrelevant to business and is recognized as a chat, a \emph{chat engine} will be employed to provide a response. If $q$ is not a chat, a pre-configured answer will be outputted.

%In both cases ($q$ is business relevant or not), if the retrieval model still fails to give an answer, a \emph{recommendation engine} will be employed for suggestions. Finally, $q$ will be transferred to staff service according to the tagged scenario if no suggestions are provided.
%Business rule parser: promotional activities.

%\begin{itemize}
%    \item .
%    \item . 
%    \item . 
%\end{itemize}

%\section{The Core of \emph{AliMe Assist}: Question Answering}
%\section{Question Answering}
\section{System Features}
\label{sec:approach}

\subsection{Intention Identification}
\label{sec:intention}

User intention in \emph{AliMe Assist} is classified into three categories: (1) requesting for assistance, e.g., ``I want to book a flight ticket''; (2) asking for information or solution, e.g., ``how to find back my password?''; (3) chatting (e.g., ``I am unhappy''). Each category is further specialized according to supported biz scenarios, e.g., assistance service supports flight booking, mobile recharge, etc.

\begin{figure}[H] %[H]
    \centering
    \vspace{-0.4cm}
    \includegraphics[width=0.95\columnwidth]{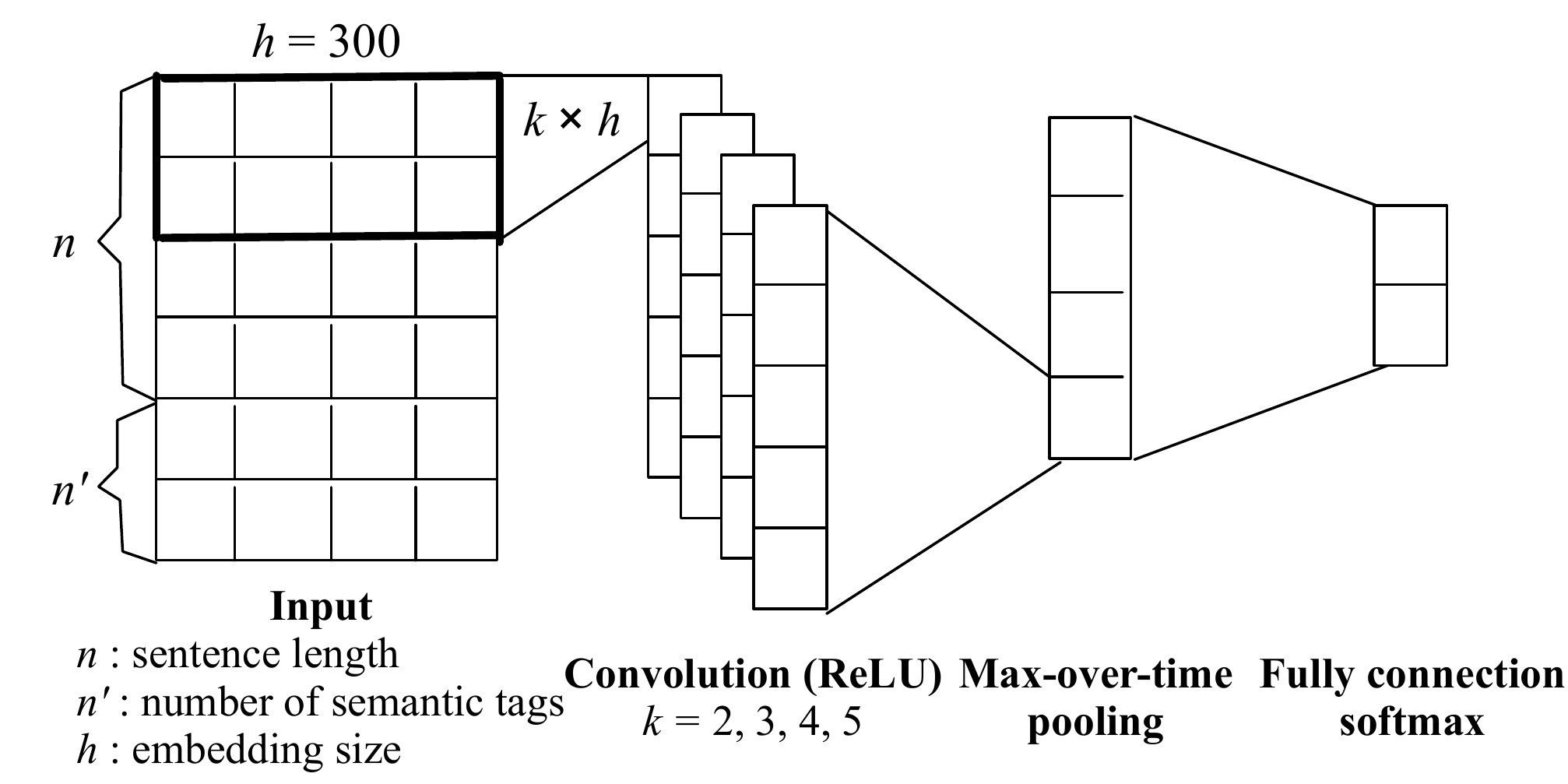}
    \vspace{-0.3cm}
    \caption{Intention classification with CNN}
    \vspace{-0.4cm}
    \label{fig:cnn}
\end{figure}

The two components, \emph{business rule parser} and \emph{intention classifier}, work together in identifying the intention of each customer question. Our business rule parser is built on hundreds of thousands of patterns obtained from frequent itemset mining and a \emph{trie}-based pattern matcher. Our intention classifier is built on CNN~\cite{kim2014convolutional}, as shown in Fig.~\ref{fig:cnn}. 

The input of our convolutional intention model is the embedding of each word of a given question $q$, followed by that of the semantic tags identified from $q$ and its previous question (i.e., its context). The embeddings are pre-trained using FastText~\cite{bojanowski2016enriching} and further fine tuned in the CNN model. Our experiment data includes 40 intention scenarios and 67,797 instances (47,455 for training and 20,342 for testing). With semantic tags, our CNN model is able to achieve a precision of 89.91\%, which is 0.91 percent higher than that without them (89\%). Both models are much better than our baseline one (an ensemble of SVM and MaxEnt, 82.71\%).

%Semantics tags are processed as follows: for each question, the identified tags, and the remaining text as a whole, are treated as individual words and trained together with the segmented words. 
%We choose one layer convolution-pooling CNN because of two practical reasons: (1) with CNN, the capture of context information (words before and after the target one) is able to achieve a ``good enough'' result in our case; (2) it is able to satisfy our performance requirement  for an average QPS (query per second) of 200 in our current setting. Recurrent NN (e.g., LSTM) is able to achieve better result, but is not practically scalable in our industrial setting. 

We choose one layer convolution-pooling CNN instead of Recurrent NN (e.g., LSTM) because: (1) CNN can also capture the context information of words in text (before and after) and is able to achieve a ``good enough'' result in our case; (2) it is efficient and is able to support a QPS (query per second) around 200 in our industrial setting. More convolution-pooling layers or RNN is able to achieve better results, but has a poorer scalability. 

\subsection{Task-Oriented Assistance Service}
\label{sec:task_qa}
%Task-oriented life assistance scenarios such as flight booking, flowers ordering and mobile recharge often require customers to provide information for predefined attributes or slots. For example, to book a flight ticket, one needs to provide the departure, the destination and the date. 

%Life assistance scenarios often require customers to provide information for certain attributes or slots. For example, to book a flight ticket, one needs to provide the departure, the destination and the date. Our solution to such task-oriented QA is to first define a schema that specifies the mandatory and optional slots for a specific task, and then use \emph{slot-filling} to extract from customer inputs and fill in values for the predefined slots. Our \emph{slot-filling} engine is mainly built on dictionaries and patterns, and is able to identify fifteen kinds of attributes, e.g., product, location and date. If any mandatory information is missing, \emph{AliMe Assist} will ask back to customers. Once all required information is collected, it will request the corresponding third-party service to complete the task.
%specific

Assistance service scenarios often require customers to provide information for certain attributes/slots in order to finish a specific task. For example, to book a flight ticket, one needs to provide the departure, the destination and the date. Our solution to such task-oriented QA is to first define a schema that specifies the mandatory and optional slots for a task, and then use \emph{slot-filling} to extract from customer inputs and fill in values for the predefined slots. Our \emph{slot-filling} engine is mainly built on dictionaries and patterns, and is able to identify fifteen kinds of attributes, e.g., product, location and date. \emph{AliMe Assist} will ask customers for mandatory information, and request third-party services to complete the task.

\subsection{Knowledge-Oriented Customer Service}
\label{sec:kn_qa}

%Knowledge graph (Entity-Relation).
%Semantic normalization.
%Information Retrieval.
%Our statistics show that 
%In \emph{AliMe Assist}, x\% of customer questions are looking for information or solutions. This kind of questions usually require precise answers (unlike chatting where relevant answers are also acceptable), and probably need to reason over knowledge items (e.g., a knowledge item that describes how to find back password can be useful if there is no specific knowledge about retrieving login password). We employ \emph{knowledge graph} for handling these knowledge-oriented questions.

%In \emph{AliMe Assist}, x\% of customer questions are looking for information/solutions. 

In \emph{AliMe Assist}, customer questions looking for information/solutions need to be addressed as precise as possible. We employ \emph{knowledge graph} to tackle such knowledge-oriented questions.

\begin{figure}[h]
    \centering
    \vspace{-0.3cm}
    \includegraphics[width=0.9\columnwidth]{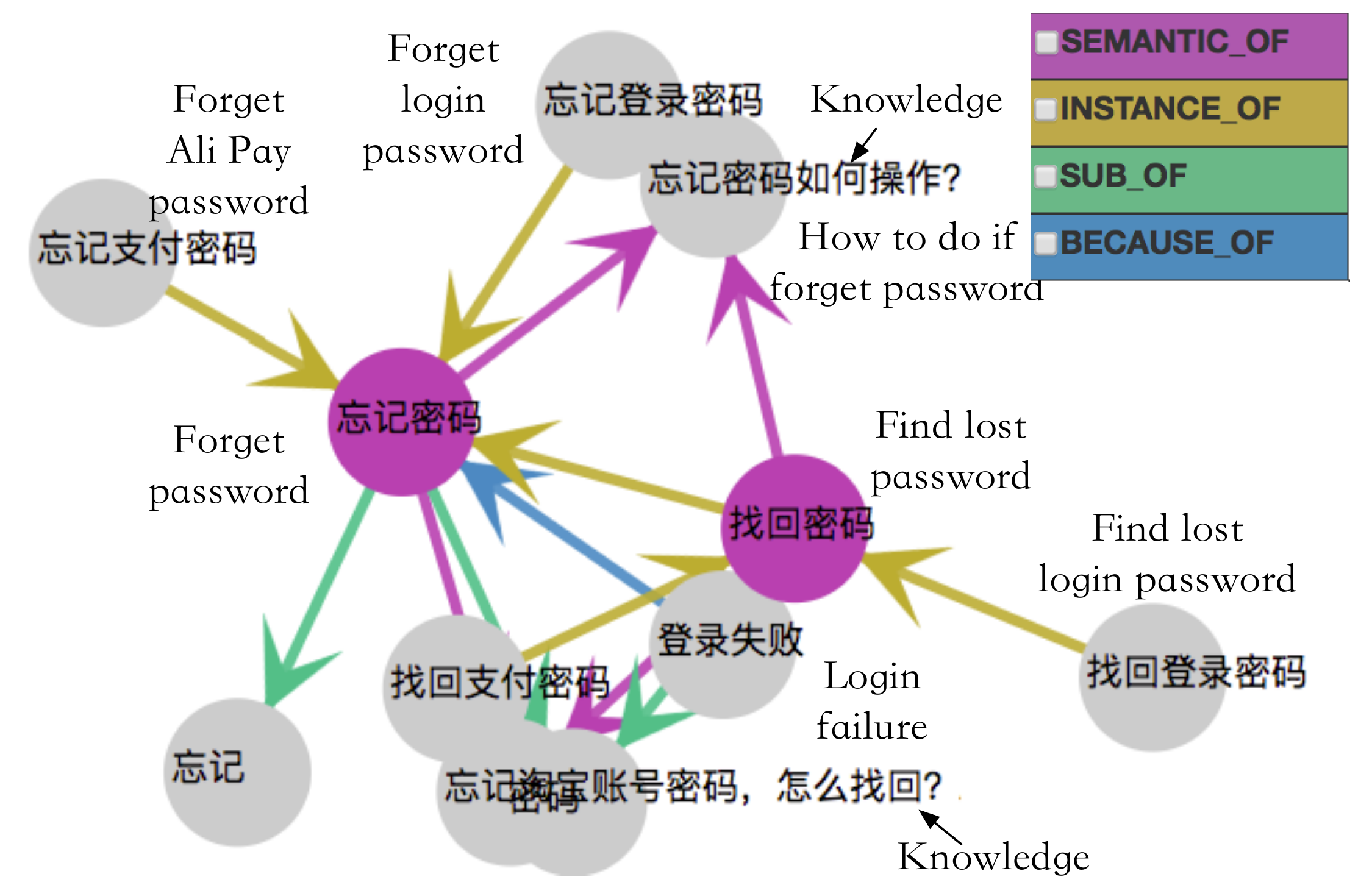}  %knowledge_graph
    \vspace{-0.3cm}
    \caption{An excerpt of a knowledge graph}
    \vspace{-0.3cm}
    \label{fig:knowledge_graph}
\end{figure}

The building blocks of our knowledge graph are \emph{entities} and \emph{relations}. We first extract candidate nouns and verbs from natural language knowledge items through word segmentation, part-of-speech (POS) tagging and tf-idf filtering, and use them to construct high-order (a combination of several, usually two) entities based on \emph{mutual information}. We then have business analysts review the entities, and design relations to build a hierarchical structure. Finally, we adopt Neo4j~\footnote{https://neo4j.com/} as our query engine. Our knowledge graph includes several thousands of entities and a fixed set of relations, and supports short (one- or two-) hop reasoning. For example, as shown in Fig.~\ref{fig:knowledge_graph}, if one asks about ``how to find my lost login password'' which does not have an associated knowledge item, we can provide a generalized but also useful answer by returning the knowledge item associated with its parent node ``find lost password''. 

%we can offer a more general but also useful answer by returning the knowledge item associated with its parent node ``find back password''. 

%~\footnote{https://neo4j.com/} 
%~\footnote{We https://github.com/lionsoul2014/jcseg}
%The building blocks of our knowledge graph are \emph{entities} and \emph{relations}. We first extract candidate nouns and verbs from natural language knowledge items (including titles and contents) through word segmentation, POS tagging and tf-idf filtering. We then have business analysts go through these candidates, pick up entities and relations, and manually build a hierarchical structure. Finally, we use Neo4j as our knowledge graph engine to empower our knowledge graph with reasoning power. For example, as in Fig.~\ref{fig:knowledge_graph}, if one asks about ``how to find back login password'' which does not have an associated knowledge item, we can offer a more general but also useful answer by returning the knowledge item associated with the parent node, e.g., ``find back password''. 

In practice, \emph{semantic normalization} -- mapping different kinds of utterances to a semantically equivalent entity in the knowledge graph -- is of key importance as customer questions are often highly diversified. We tackle this problem through {utterance diversification} and {pattern mining}. In the first stage, given a set of knowledge items $K$, we identify its diversified utterances through finding similar answers $K'$ from historical chat logs (between customers and service staff) and taking the corresponding questions for $K'$. The similarity calculation algorithm is designed based on sentence embedding~\cite{le2014distributed}, which helps to capture semantic similarity, and implemented using Map-Reduce~\cite{dean2008mapreduce}. Once we have the mapping between a knowledge item and a set of diversified utterances, we are able to extract wording patterns for entities using frequent itemsets mining. These patterns will be used in the trie-based semantic parser to identify semantic tags (again, entities in knowledge graph) for each customer question.

% between questions/answers
%SimHash~\cite{manku2007detecting}
%To measure the similarity between short texts (e.g., sentences) in a big data set, we have proposed our optimized pre-selection algorithms based on traditional inverted index and SimHash~\cite{manku2007detecting}, and distributed similarity calculation algorithms based on Map-Reduce. 

%One point to be noted is that the textual feature of knowledge items is enriched with with uni-grams, bi-grams and synonyms (besides words) in our  similarity calculation stage. 

%we mine frequent item sets group similar sentences (questions or knowledge items) into a cluster through our optimized hierarchical clustering algorithm. 
%we first maintain a synonyms Table through word mining (plus manual review) based on word embedding. 
%Knowledge graph is designed for high precision knowledge-oriented QA at the cost of certain recall decrease. 
 
 %but also leads to loss of recall.  traditional
%Compared with traditional information retrieval (IR) model, our knowledge graph approach increases accuracy by 10\%. To improve recall, retrieval model is used as a complement to process those questions not addressed by the knowledge graph engine.

Compared with traditional information retrieval (IR) model, our knowledge graph approach increases accuracy by 10\%. Since our knowledge graph covers only frequently asked (i.e., a small subset of) knowledge items, retrieval model is used as a complement to improve recall.

%Meanwhile, to boost recall, questions not addressed by the knowledge graph engine will be fed into a retrieval model. To ensure precision, the similarity threshold is set to relatively high. 

%Our retrieval model follows a typical process: query analysis, candidates recall, similarity calculation and answer processing. 

%Meanwhile, to boost recall, retrieval model is also employed to handle the questions not addressed by the knowledge graph engine. Our retrieval model follows a typical process: query analysis, candidates recall, similarity calculation and answer processing. To ensure precision, the similarity threshold is set to relatively high. 

%Currently, our\ retrieval model takes around 5\% - 10\% of the questions in total. 
%those questions that cannot be handled by knowledge graph.
%\subsubsection{Chat-Oriented QA}

\subsection{Chatting Service}
\label{sec:chat_qa}

In \emph{AliMe Assist}, the majority of customer questions is business-related, but also around 5\% of them is chat-oriented (several hundreds of thousands in number). To offer better user experience, we build an open-domain chatbot engine, where we propose a hybrid approach that uses an attentive \emph{Sequence-to-Sequence} (Seq2Seq) model to optimize the joint results of IR models~\cite{dochat} and Seq2Seq generation models~\cite{le-nips,le-workshop}. 

%\cite{ji-ir14,dochat}
%~\cite{le-nips,le-workshop}
 
%it is necessary to 
%, which outperforms both retrieval and Seq2Seq model
% We propose a hybrid approach that integrates Information Retrieval (IR) model with Sequence-to-Sequence (Seq2Seq) model. 
%bahdanau-attention,
%We propose a hybrid approach that integrates Information Retrieval (IR) model~\cite{ji-ir14,dochat} with Sequence-to-Sequence (Seq2Seq) generation model~\cite{le-nips,le-workshop}. Specifically, we use an attentive Seq2Seq rerank model~\cite{bahdanau-attention} to optimize the joint results. 

\begin{figure}[h] %[H]
    \centering
    \vspace{-0.2cm}
    \includegraphics[width=1.0\columnwidth]{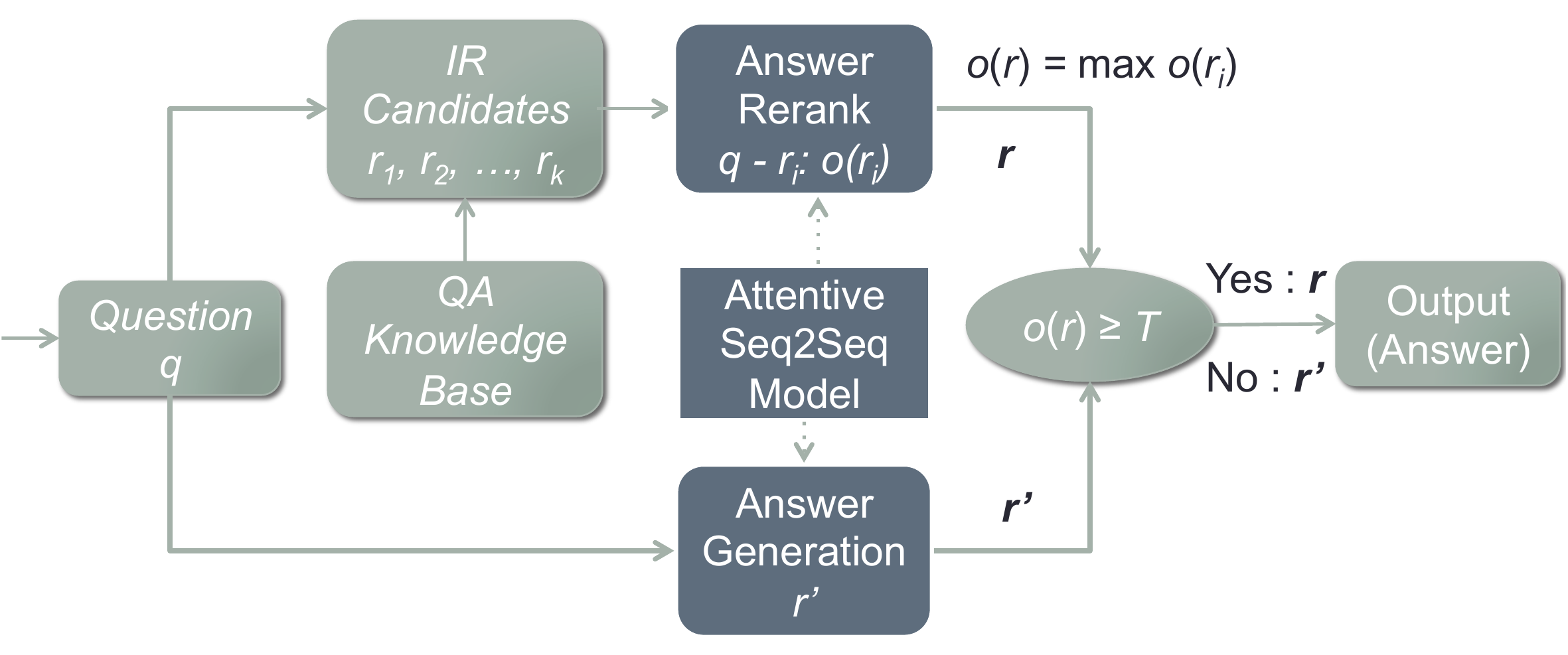}
    \vspace{-0.8cm}
    \caption{An overview of our hybrid chat approach}
    \vspace{-0.3cm}
    \label{fig:chat_hybrid}
\end{figure}

The overview of our approach is shown in Fig.~\ref{fig:chat_hybrid}. First, a retrieval model is used to collect candidate answers. Second, an attentive Seq2Seq model is employed to rerank the candidates by giving each candidate a confidence score. Top candidate with a confidence score that exceeds a given threshold can be used as an answer. If it is lower than the threshold, the generated answer by Seq2Seq model will be taken as the output. Interested readers can refer to~\cite{minghui17alimechat} for details.
% (these candidates have been ranked by retrieval model itself)
%(0.19, obtained from empirical studies) 
%from a semantic view 

%Two techniques, namely \emph{Information Retrieval} and \emph{Sequence-to-Sequence} (Seq2Seq) model, are used for chat-oriented QA. Individually, retrieval model can only obtain answers in a pre-established knowledge base, while Seq2Seq model is able to generate answers out of the box but the answers sometimes can be inconsistent or meaningless. 

%Our approach leverages the advantages of both IR and Seq2Seq,

%Our approach alleviates the flaws of both IR and Seq2Seq generation models: the former can only handle questions that are close to those in a QA knowledge base, while the latter may generate inconsistent or meaningless answers. 
%top-1 accuracy
Extensive experiments show that our approach significantly outperforms IR and generation models: in a test with 600 questions, the $P_{top_1}$ (whether the top-1 candidate is acceptable) of our approach, IR and Seq2Seq generation are 60.01\%, 47.11\% and 52.02\%; an online A/B test (2136 questions, 1047 by the hybrid approach, 1089 by IR) also confirms that our hybrid approach performs much better than IR ($P_{top_1}$: 60.36\% vs. 40.86\%). 

%\fq{add citation}.
%We conducted a set of experiments to evaluate the proposed approach. Results show that our hybrid approach outperforms both IR and Seq2Seq generation model (the top-1 accuracy of the three approaches are 60.01\%, 47.11\% and 52.02\%, respectively).

%As we can see from the result of an experiment in Fig~\ref{fig:chat_results}, the hybrid approach has the best top-1 accuracy ($P_top_1$): with a confidence score $T = 0.19$, the top-1 accuracy achieves 60.01\%. Here, questions with a score higher than 0.19 (the left of the dashed line), are answered using rerank, and the rest is handled by generation. One may note that a narrowly higher ($P_top_1$) can be achieved if a higher threshold is used (e.g., 0.48), or, put differently, rerank less and generate more. The lower threshold is used because of the uncontrollability and poor interpretability of Seq2Seq generation: with an elegant decrease at the ($P_top_1$) , it is able to gain more controllability and interpretability. 

%\begin{figure}[H]
%    \centering
%    \vspace{-0.2cm}
%    \includegraphics[width=1.0\columnwidth]{chat_results}
%    \vspace{-0.8cm}
%    \caption{The results of our hybrid approach for chat-oriented QA}
%    \vspace{-0.2cm}
%    \label{fig:chat_results}
%\end{figure}

\section{Demonstration}
\label{sec:tool}
%\begin{figure}[h]
%In this section, we demonstrate the three key services of \emph{AliMe Assist}.
%We demonstrate the key features of \emph{AliMe Assist} -- operation assistance, after-sales service and chatting -- in Fig~\ref{fig:book_flight_demo}.
%and image
% by registering a Taobao account or using your mobile phone for identify validation

%We launch \emph{AliMe Assist} for a real-world industrial intelligent customer service assistant that offers the three kinds of services: operation assistance, after-sales service and chatting service~\footnote{Interested readers can access the full version \emph{AliMe Assist} through the Taobao App by following the path ``我的淘宝(My Taobao) $\to$ 我的小蜜(My Ali Me)'', or the web version via https://consumerservice.taobao.com/online-help (text only, operation assistance not supported).}. 

%We launch \emph{AliMe Assist} for a real-world industrial intelligent assistant that offers the three kinds of services: life assistance, customer service and chatting service~\footnote{Interested readers can access \emph{AliMe Assist} through the Taobao App by following ``我的淘宝(My Taobao) $\to$ 我的小蜜(My Ali Me)'', or the web version via https://consumerservice.taobao.com/online-help (text only, life assistance not enabled).}. 

We launch \emph{AliMe Assist} for a real-world industrial intelligent assistant~\footnote{Interested readers can access \emph{AliMe Assist} through the Taobao App by following ``My Taobao $\to$ My AliMe'', or the web version via https: //consumerserv-ice.taobao.com/online-help (text only, assistance service not enabled).}, which currently serves millions of customer questions per day and is able to address 85\% of them. 

\begin{figure*}[h]
    \centering
    \vspace{-0.2cm}
    \includegraphics[width=0.81\textwidth]{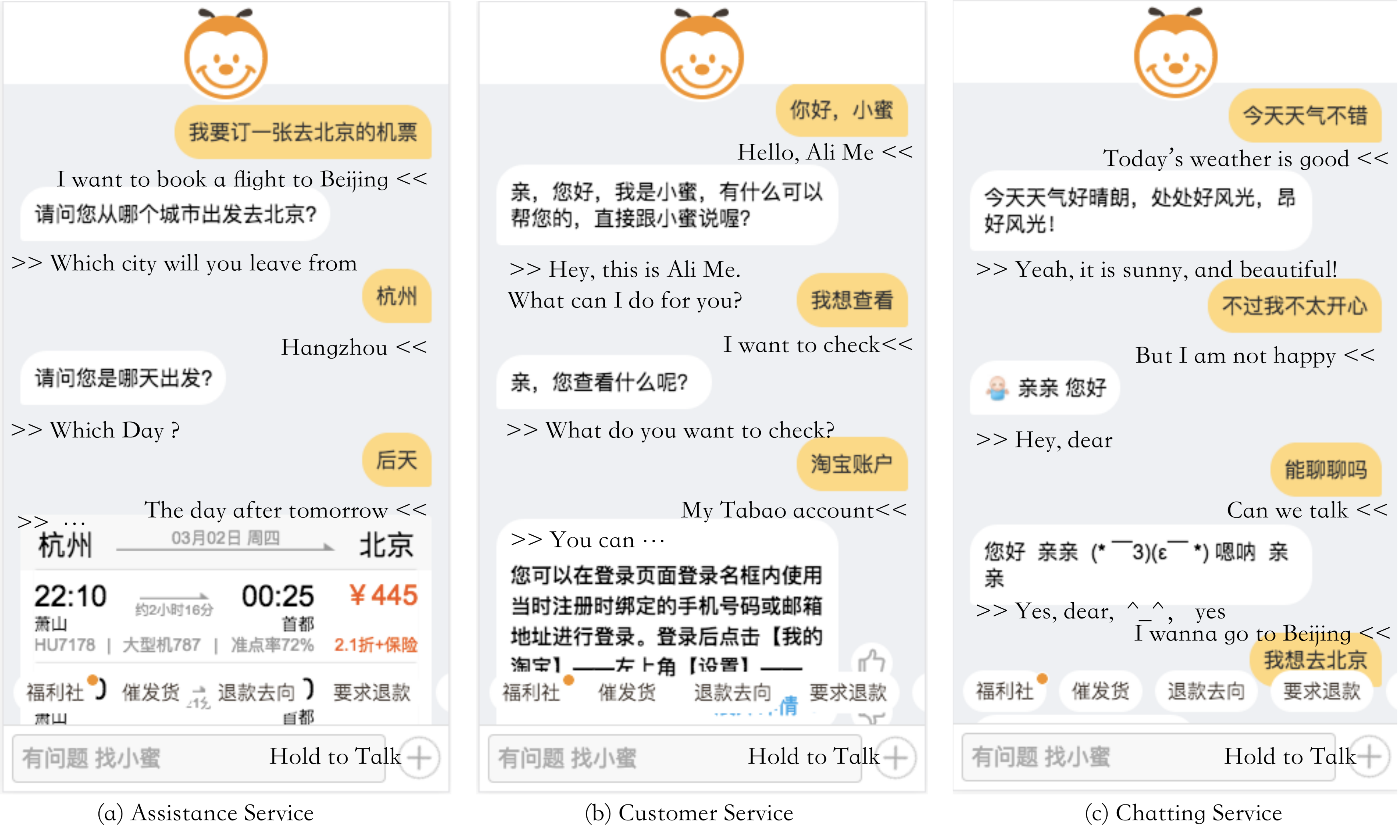}
    \vspace{-0.4cm}
    \caption{A demonstration of \emph{AliMe Assist}}
    \vspace{-0.4cm}
    \label{fig:demo}
%\end{figure}
\end{figure*}

%(e.g., first the departure and then the date)
% The last Fig~\ref{fig:demo} (c) shows how \emph{AliMe Assist} chat with users when they come just for fun or do not have a clear purpose. 
We demonstrate the key features, namely assistance service, customer service and chatting service, through three realistic scenarios in Fig~\ref{fig:demo}. Fig~\ref{fig:demo} (a) illustrates how \emph{AliMe Assist} helps users to book a flight ticket through asking for needed information step by step, and then calling Ali Trip, a travel agency of Alibaba Group, to complete the task. Fig~\ref{fig:demo} (b) demonstrates how \emph{AliMe Assist} helps to address customer service questions by combining the previous question ``I want to check'' and the current one ``Taobao account'', and then employing the knowledge graph engine to provide an answer. Fig~\ref{fig:demo} (c) shows a chat conversation between \emph{AliMe Assist} and customers. These three scenarios together demonstrate the intention identification feature: given a question, \emph{AliMe Assist} is able to identify its intention, and distribute it to the corresponding service (e.g., assistance service) and scenario (e.g., booking a flight). 

\section{Related Work}
\label{sec:relwork}

%Recall that our AliMe Assist is built on three techniques: task-oriented, knowledge-oriented, and chat-oriented QA. We organize the related work accordingly.
In this section, we review related work on task-, knowledge-, and chat-oriented QA, which are closely related to the techniques employed in \emph{AliMe Assist}.

%\cite{end-end-lstm}
%\cite{MrksicSTGSVWY15}
%~\cite{matthew,wang-SIGDIAL}
\textbf{Task-oriented QA.} This belongs to closed domain QA. Typically people use rule- or template- based methods~\cite{HsienWen16}, and dialog state tracking~\cite{matthew}. Our slot-filling method is similar to the template-based method.

%Often, rule- or template- based method~\cite{HsienWen16}, and dialog state tracking~\cite{matthew} are used for this closed domain QA. Our slot-filling method is similar to the template-based method.

%

%\textbf{Knowledge-oriented QA.} This technique can be used in both closed- and open- domain. Usually data-driven techniques such as Information Retrieval (IR) methods are used. IR based techniques mainly focus on finding the nearest question(s) from a QA knowledge base for an input question, e.g.,  \cite{isbell-ir00},  ~\cite{ji-ir14},~\cite{dochat}. A recent work \cite{yan-ir-16} has tried a neural network based method for matching. Usually, IR based models have difficulty in handling long-tail questions. Different from these methods, our engine uses both knowledge graph and IR techniques. The former is used to handle those popular customer questions specific to E-commerce domain. And the latter is used to handle those unsolved questions. To build a knowledge graph, we use information extraction techniques~\cite{yao2014information,Jiang2012}.

%\cite{isbell-ir00},  ~\cite{ji-ir14},
\textbf{Knowledge-oriented QA.} This technique can be used in both closed and open domain. Usually, data-driven IR model is used here. The idea is to find the nearest question(s) from a QA knowledge base for each input question, e.g.,~\cite{dochat}. A recent work \cite{yan-ir-16} has tried a neural network based method for matching. Being different from these methods, we uses both knowledge graph and IR: the former is designed for high precision QA in E-commerce and the latter is used as a complement.

% In general, IR based models have difficulty in handling long-tail questions. 
%To build a knowledge graph, we use information extraction techniques~\cite{yao2014information,Jiang2012}.

%\textbf{Chat-oriented QA.} Usually this technique is for open-domain. Commonly used techniques include IR model~\cite{ji-ir14,dochat} and generation model~\cite{bahdanau-attention,le-nips,le-workshop}. Given a question, the former retrieves the nearest question in a Question-Answer (QA) knowledge base and takes the paired answer, the latter generates an answer based on a pre-trained Seq2Seq model. Often, IR models fail to handle long-tail questions that are not close to those in a QA base, and generation models may generate inconsistent or meaningless answers~\cite{li-diversity-naacl,SerbanSBCP16}. Another recent combinational approach~\cite{combined-work} uses an IR model to rerank the union of retrieved and generated answers. Our work differs from ~\cite{combined-work} in that we use attentive Seq2Seq rerank approach to integrates both IR and generation models.

%Given a question, the former retrieves the nearest question in a Question-Answer (QA) knowledge base and takes the paired answer, the latter generates an answer based on a pre-trained Seq2Seq model. 

%\cite{ji-ir14,dochat} 
%\cite{bahdanau-attention,le-nips,le-workshop}
%\cite{li-diversity-naacl,SerbanSBCP16}
\textbf{Chat-oriented QA.} Commonly used methods for this open domain QA include IR~\cite{dochat} and generation models~\cite{bahdanau-attention,le-nips}. Our approach alleviates the flaws of both IR and Seq2Seq generation models: the former can only handle questions that are close to those in a QA knowledge base, while the latter may generate inconsistent or meaningless answers~\cite{SerbanSBCP16}. Another recent combinational approach~\cite{combined-work} uses an IR model to rerank the union of retrieved and generated answers. Our work differs from it in that we use an attentive Seq2Seq rerank approach to optimize the joints results of IR and generation models.
%~\cite{combined-work}

%\vspace{-0.5cm}
\section{Conclusion}
\label{sec:conclusion}
In this paper, we present \emph{AliMe Assist}, an intelligent assistant that is designed for creating an innovative E-commerce experience. We launch \emph{AliMe Assist} for a real-world application that offers customers with assistance service, customer service and chatting service, and currently handles millions of customer questions per day. 

As for future work, several points will be further explored to improve our assistant. For example, strengthening context based multi-round interaction~\cite{wu2016sequential}, offering shopping guidance based on Reinforcement Learning (RL), empowering \emph{AliMe Assist} with the capability to ``read'' images through image recognition, etc.

\vspace{-0.1cm}
\bibliographystyle{ACM-Reference-Format}
\bibliography{cikm2017} 

\end{document}